\pdfoutput=1

\documentclass[11pt]{article}

\usepackage{emnlp2021}

\usepackage{times}
\usepackage{latexsym}

\usepackage[T1]{fontenc}

\usepackage[utf8]{inputenc}

\usepackage{microtype}

\usepackage{soul}
\usepackage{url}
\usepackage[utf8]{inputenc}
\usepackage{graphicx}
\usepackage{amsmath}
\usepackage{amsthm}
\usepackage{booktabs}
\usepackage{algorithm}
\usepackage{algorithmic}

\usepackage{amssymb}
\usepackage{color}
\usepackage[switch]{lineno}
\usepackage{multirow}
\usepackage{subfig}
\usepackage{xspace}
\title{When Retriever-Reader Meets Scenario-Based Multiple-Choice Questions}

\author{
    Zixian Huang \and Ao Wu \and Yulin Shen \and Gong Cheng \and Yuzhong Qu \\
    State Key Laboratory for Novel Software Technology, Nanjing University, China \\
    \texttt{\{zixianhuang,awu,ylshen\}@smail.nju.edu.cn} \\ \texttt{\{gcheng,yzqu\}@nju.edu.cn}
}

\begin{document}
\maketitle
\begin{abstract}
Scenario-based question answering (SQA) requires retrieving and reading paragraphs from a large corpus to answer a question which is contextualized by a long scenario description. Since a scenario contains both keyphrases for retrieval and much noise, retrieval for SQA is extremely difficult. Moreover, it can hardly be supervised due to the lack of relevance labels of paragraphs for SQA. To meet the challenge, in this paper we propose a joint retriever-reader model called JEEVES where the retriever is implicitly supervised only using QA labels via a novel word weighting mechanism. JEEVES significantly outperforms a variety of strong baselines on multiple-choice questions in three SQA datasets.
\end{abstract}

\section{Introduction}
\label{sec:introduction}

Scenario-based question answering (SQA) is the task of answering a question which is contextualized by a \emph{scenario} describing a concrete case~\cite{watsonpaths},
e.g.,~predicting a judgment based on a legal scenario~\cite{DBLP:conf/acl/XuWCPWZ20}, answering a multiple-choice question in China's national college entrance examination (i.e.,~GaoKao) based on a geographical scenario~\cite{tsqa}. SQA is a challenging task as it combines the difficulties of open-domain question answering (ODQA)~\cite{DRQA,DBLP:journals/tacl/KwiatkowskiPRCP19} and machine reading comprehension (MRC)~\cite{SQuAD,RACE}. Indeed, SQA not only relies on accurately \emph{retrieving} relevant paragraphs from a large corpus, but also requires thoroughly \emph{reading} the retrieved paragraphs and the provided scenario to fuse their knowledge and infer an answer.

\begin{figure}[t!]
    \centering
    \includegraphics[width=\columnwidth]{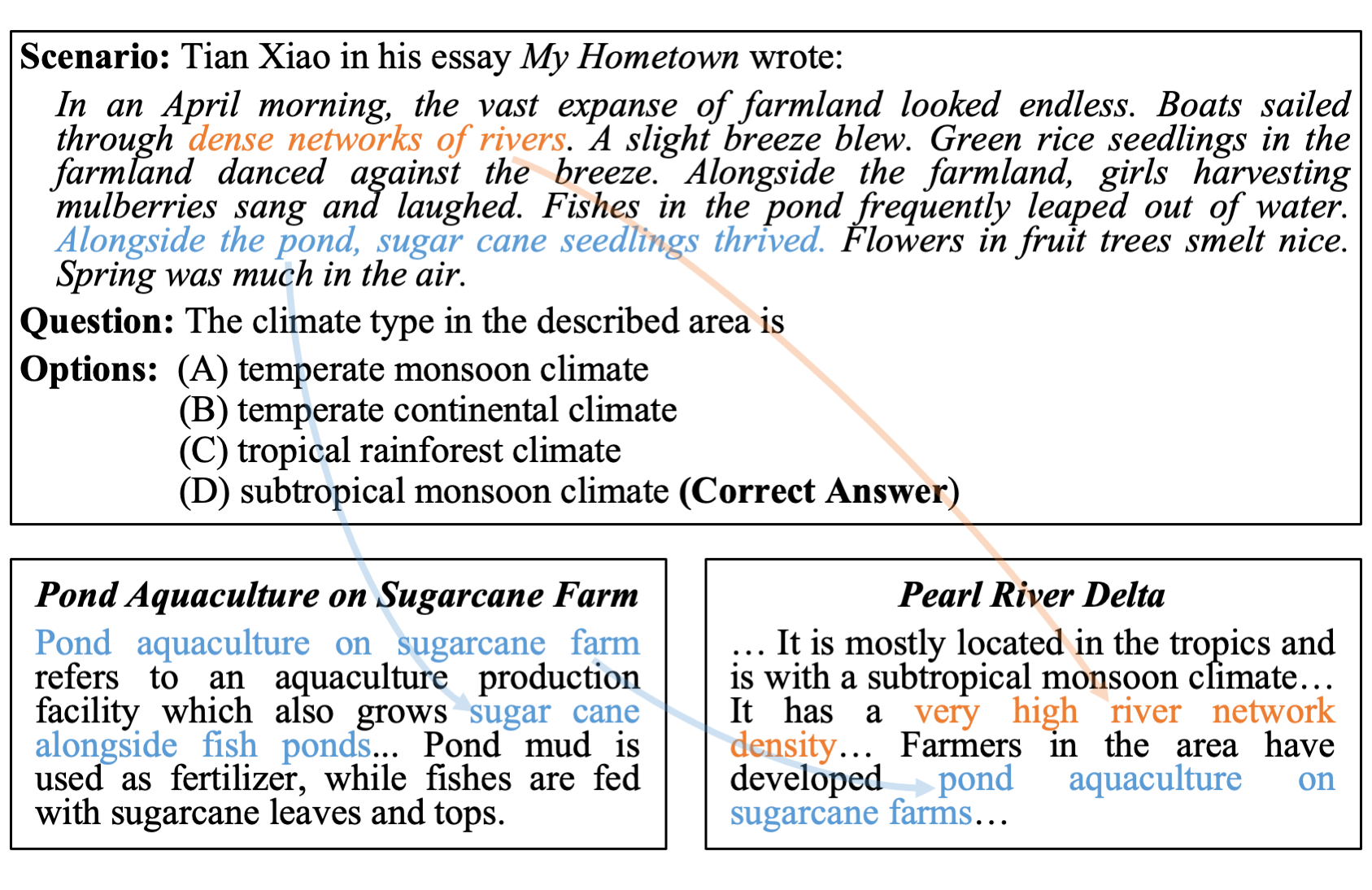}
    \caption{\textbf{Top:} a scenario and a multiple-choice question about geography; in the scenario we highlight some keyphrases for retrieval to support answering the question. \textbf{Bottom:} some supporting paragraphs in two articles retrieved from an online encyclopedia. \textbf{Arrows:} two reasoning paths that connect the area described in the scenario with a similar area in the real world.}
    \label{fig:sampleproblem}
\end{figure}

For example, Figure~\ref{fig:sampleproblem} shows a scenario and a multiple-choice question sampled from geography exams in GaoKao. For machines to answer such a question, it would be insufficient to only read the given scenario but is necessary to retrieve from a corpus (e.g.,~textbooks, Wikipedia) to acquire supporting knowledge. In this example, the scenario gives clues about the location of the described area. The keyphrases we highlight in the scenario connect the described area with the Pearl River Delta via two reasoning paths. The delta has a subtropical monsoon climate, indicating the correct answer. This multi-hop reasoning process fuses the scenario and the retrieved supporting paragraphs.

\paragraph{Research Challenge.}
SQA, similar to ODQA, is commonly solved by a retriever-reader framework. However, retrieval for SQA is fundamentally more difficult than for ODQA because a scenario typically gives a long paragraph containing both \emph{keyphrases for retrieval} and much \emph{noise}, as illustrated in Figure~\ref{fig:sampleproblem}. Hence, in SQA, the paragraphs found by unsupervised retrievers such as BM25 are often not truly useful for answering the question.
Retrieval errors have become a bottleneck in SQA~\cite{gaokao-history,GeoSQA}.

Supervised retrievers may be more powerful but they cannot directly apply to SQA due to the \emph{lack of relevance labels of paragraphs} for training. Only QA labels (i.e.,~correct answers) are available. Recent dense passage retrievers (DPR)~\cite{DPR} can be distantly supervised by QA labels, using paragraphs that contain the gold-standard answer as positives. While such distant supervision is effective for ODQA where answers are short, it could not generate high-quality positives for most questions in SQA where the gold-standard answer is sentence-long and \emph{not explicitly contained in any paragraph}. As a result, DPR shows unsatisfying performance on SQA datasets, as our experimental results will show.
How can we effectively supervise a retriever for SQA?

\paragraph{Our Approach---JEEVES.}
To meet the challenge, we propose a novel retriever-reader method called JEEVES where we employ only QA labels to \emph{implicitly} train our retriever. It is realized by learning a \emph{word weighting} for distinguishing keyphrases from noise in the scenario to improve the accuracy of retrieval and hence of the downstream QA which is supervised. Since our retriever and reader are both supervised by QA labels, we train them jointly to further improve the overall accuracy.

We summarize our contributions as follows.
\begin{itemize}
    \item We propose to implicitly supervise a retriever for answering multiple-choice questions in SQA by learning a word weighting.
    \item We present JEEVES, a new end-to-end model for SQA where the retriever and the reader are jointly trained only with QA labels.
    \item JEEVES significantly outperforms state-of-the-art methods on three SQA datasets, including a new dataset we construct.
\end{itemize}


Our code and data are on GitHub: \url{https://github.com/nju-websoft/Jeeves-GKMC}.

\paragraph{Outline.}
We discuss related work in Section~\ref{sec:rw}, describe JEEVES in Section~\ref{sec:model}, present experiments in Section~\ref{sec:experiments}, and conclude the paper in Section~\ref{sec:conclusion}.
\section{Related Work}
\label{sec:rw}

\paragraph{SQA}
has found application in many domains~\cite{watsonpaths,legalemnlp18,DBLP:conf/acl/ChalkidisAA19,DBLP:conf/ijcai/YangJZL19,DBLP:conf/acl/XuWCPWZ20,gaokao-history,GeoSQA,tsqa}. Among others, \citet{GeoSQA} published the GeoSQA dataset containing multiple-choice questions in high-school geography exams, and they evaluated a variety of methods, mostly adopting a retriever-reader framework. All the tested methods performed close to random guess. The authors pointed out that a major cause of error is imprecise retrieval due to the long description of a scenario which is critical to retrieval but noisy.

JEEVES mainly aims at addressing this issue and it outperformed state-of-the-art methods on three SQA datasets in our experiments.

\paragraph{Retrievers}
for SQA as well as ODQA are mainly unsupervised (e.g.,~BM25) due to the lack of relevance labels of paragraphs.
Although word matching based retrievers are still useful~\cite{ORQA}, they could not effectively exploit long and noisy scenario descriptions in SQA and became a bottleneck~\cite{gaokao-history,GeoSQA}. Supervised retrievers such as rerankers~\cite{rerank} and dense retrievers~\cite{ORQA, DPR,ColBERT} need labeled paragraphs for training. While dense retrievers have been effectively trained with labels from distant supervision on ODQA datasets, in our experiments they were less effective on SQA datasets such as GeoSQA where answers (i.e.,~options of multiple-choice questions) are sentence-long and often not explicitly contained in any paragraph. As a result, distant supervision by exact matching failed to generate positives, while approximate matching generated false positives.

Another way to handle the lack of labels is by transfer learning. For example, AR~\cite{AR} trains its retriever with labeled paragraphs from another similar dataset. However, we are not aware of any datasets that contain labeled paragraphs and are sufficiently similar to SQA datasets like GeoSQA. Training with labeled paragraphs from a very different dataset would influence the accuracy of the learned retriever on the target dataset. Indeed, AR showed unsatisfying performance on SQA datasets in our experiments.

JEEVES does not rely on distant supervision but only employs QA labels in the target SQA dataset to implicitly yet effectively train a retriever.


\paragraph{Readers} have been extensively studied in MRC research. State-of-the-art methods use pre-trained language models (PLMs)~\cite{MMM,DCMN+,SG-Net,DHC}. Some reader-retriever frameworks for ODQA jointly train their retriever and reader~\cite{R3,ORQA}. JEEVES incorporates a reader which also builds on PLMs and fuses the knowledge in multiple paragraphs. JEEVES also performs joint training. These design choices showed effectiveness in our ablation study, although they are not included in our main research contributions.

\section{Approach}
\label{sec:model}

\subsection{Task Definition}

We focus on scenario-based multiple-choice questions. Given a scenario description~$S$, a question~$Q$ has a set of $m$~options $\mathbf{O}=\{O_1, \ldots, O_m\}$ where exactly one option is the correct answer. For each option $O_i \in \mathbf{O}$, we refer to the concatenation of~$S$, $Q$, and~$O_i$ as an enriched option~$\hat{O}_i$. Observe that to answer~$Q$, we commonly need to fuse the description in~$S$ with knowledge in some supporting paragraphs retrieved from a corpus~$\mathbf{P}$.

\subsection{Overview of JEEVES}

\begin{figure}[t!]
    \centering
    \includegraphics[width=\columnwidth]{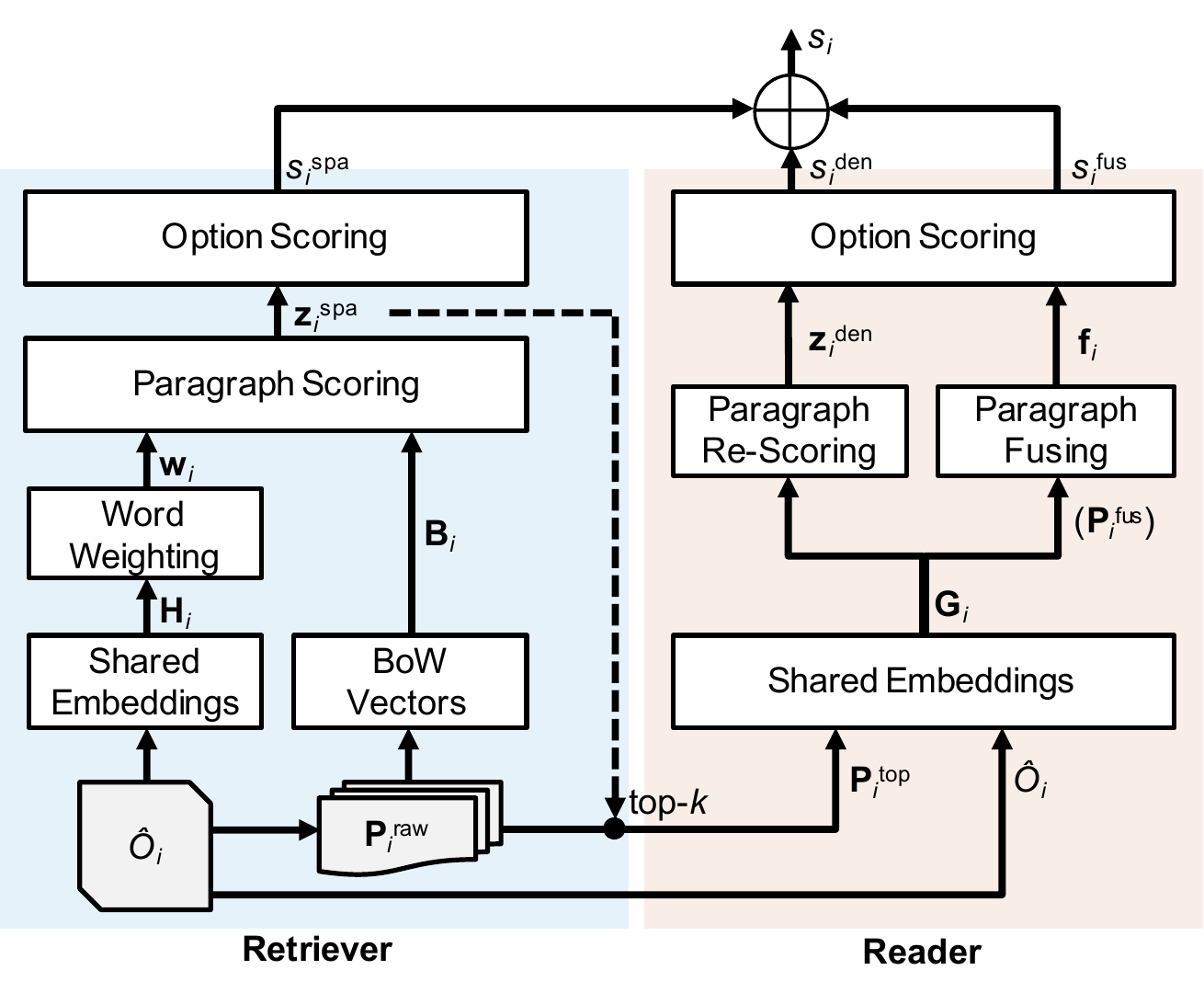}
    \caption{Overview of JEEVES.}
    \label{fig:model}
\end{figure}

Figure~\ref{fig:model} presents an overview of JEEVES,
which adopts a retriever-reader framework.

\paragraph{Retriever.}
For each enriched option~$\hat{O}_i$, we consider all the potentially relevant paragraphs~$\mathbf{P}^\text{raw}_i$ in the corpus. Since $\mathbf{P}^\text{raw}_i$~may have a very large size, for the paragraphs in~$\mathbf{P}^\text{raw}_i$ we only generate their sparse representations~$\mathbf{B}_i$ over the words in~$\hat{O}_i$, but we weight these words based on the dense representation~$\mathbf{H}_i$ of~$\hat{O}_i$. We combine sparse representations~$\mathbf{B}_i$ and word weights~$\mathbf{w}_i$ to score each paragraph in~$\mathbf{P}^\text{raw}_i$ and retrieve $k$~top-ranked paragraphs~$\mathbf{P}^\text{top}_i$ to be fed into the reader. In particular, we leverage the computed paragraph scores~$\mathbf{z}^\text{spa}_i$ to compute a score~$s^\text{spa}_i$ for option~$O_i$, which in the training phase is compared with QA labels to implicitly supervise the retriever.

\paragraph{Reader.}
Since $\mathbf{P}^\text{top}_i$~has a small size of~$k$, for the paragraphs in~$\mathbf{P}^\text{top}_i$ we can afford to generate their dense representations~$\mathbf{G}_i$. We use~$\mathbf{G}_i$ to re-score each paragraph in~$\mathbf{P}^\text{top}_i$, and we choose $k'$~top-ranked paragraphs~$\mathbf{P}^\text{fus}_i$ to fuse their dense representations into~$\mathbf{f}_i$ and compute a score~$s^\text{fus}_i$ for option~$O_i$. We also leverage the re-computed paragraph scores~$\mathbf{z}^\text{den}_i$ to compute a score~$s^\text{den}_i$ for~$O_i$. Finally, $s^\text{spa}_i$, $s^\text{den}_i$, and~$s^\text{fus}_i$ are combined to score~$O_i$.

\subsection{Retriever}

Our retriever for SQA is implicitly supervised by QA labels via a novel word weighting mechanism.

\paragraph{Paragraph Representation (BoW Vectors).}
For each enriched option~$\hat{O}_i$, from the corpus~$\mathbf{P}$ we take all the potentially relevant paragraphs~$\mathbf{P}^\text{raw}_i$: those containing any non-stopword in~$\hat{O}_i$. Since $\mathbf{P}^\text{raw}_i$~may have a large size, for each paragraph $p_l \in \mathbf{P}^\text{raw}_i$ we inexpensively generate its sparse representation~$\mathbf{b}_l$ capturing its lexical features. Specifically, we adopt a bag-of-words (BoW) model. Let~$n_i$ be the number of unique words in~$\hat{O}_i$. We define~$\mathbf{b}_l$ as an $n_i$-dimensional vector where each value is the BM25 score of~$p_l$ given the corresponding word in~$\hat{O}_i$ as a query.
We pack together such vectors for all the paragraphs in~$\mathbf{P}^\text{raw}_i$ into a matrix~$\mathbf{B}_i$:
\begin{equation}
\small
    \mathbf{B}_i = [\mathbf{b}_1; \mathbf{b}_2; \ldots] \,.
\end{equation}

\paragraph{Option Representation (Shared Embeddings).}
For~$\hat{O}_i$ we generate its dense representation capturing its semantic features by feeding it into a PLM~\cite{WWM,ERNIE} which will be shared with our reader:
\begin{equation}
\label{eq:enc}
\small
      \text{PLM}(\text{[CLS] $S$ [SEP] $Q$ [SEP] $O_i$ [SEP]}) \,.
\end{equation}
\noindent Recall that $\hat{O}_i$~contains $n_i$~unique words. If a word appears in multiple positions in~$\hat{O}_i$, we will aggregate their dense representations in Eq.~(\ref{eq:enc}) by max pooling. We pack together the aggregate vectors for all the $n_i$~unique words in~$\hat{O}_i$ into a matrix~$\mathbf{H}_i$:
\begin{equation}
\small
    \mathbf{H}_i = [\mathbf{h}_1; \ldots; \mathbf{h}_{n_i}] \,.
\end{equation}

\paragraph{Word Weighting.}
Some words in~$\hat{O}_i$ are keyphrases for retrieval while others are noise. We learn to weight each word representing its salience. Specifically, for all the $n_i$~unique words in~$\hat{O}_i$, we feed their dense representations~$\mathbf{H}_i$ into two dense layers (called \emph{word weighting network}) to output $n_i$~values normalized by softmax as word weights:
\begin{align}
\label{eq:weighting}
\small
\begin{split}
    \mathbf{w}_i & = \text{softmax}([w_{i,1}; \ldots; w_{i,n_i}]^\intercal) \,,~ \text{where} \\
    w_{i,j} & = \text{linear}(\text{tanh}(\text{linear}(\mathbf{h}_j))) \text{ for } 1 \leq j \leq n_i \,.
\end{split}
\end{align}

\paragraph{Paragraph Scoring.}
For all the paragraphs in~$\mathbf{P}^\text{raw}_i$, we compute weighted sums over their sparse representations~$\mathbf{B}_i$
to obtain their scores~$\mathbf{z}^\text{spa}_i$:
\begin{equation}
\label{eq:retrieve}
\small
    \mathbf{z}^\text{spa}_i = \mathbf{B}_i^\intercal \times \mathbf{w}_i \,.
\end{equation}
\noindent We retrieve $k$~top-ranked paragraphs~$\mathbf{P}^\text{top}_i \subseteq \mathbf{P}^\text{raw}_i$ according to~$\mathbf{z}^\text{spa}_i$ and feed ~$\mathbf{P}^\text{top}_i$ into the reader.

\paragraph{Option Scoring.}
An enrich option stated explicitly in a paragraph is likely the correct answer. Paragraph scores~$\mathbf{z}^\text{spa}_i$ represent the confidence that such a statement is found~\cite{IR-Solver}. Therefore, we leverage~$\mathbf{z}^\text{spa}_i$ to compute a score~$s^\text{spa}_i$ for option~$O_i$ by feeding~$\mathbf{z}^\text{spa}_i$ into two dense layers:
\begin{equation}
\label{eq:sri}
\small
    s^\text{spa}_i = \text{linear}(\text{tanh}(\text{linear}(\mathbf{z}^\text{spa}_i))) \,,
\end{equation}
\noindent where we trim~$\mathbf{z}^\text{spa}_i$ to its $\tau$~largest values. In this way, retrieval results~$\mathbf{P}^\text{top}_i$ are connected with QA results~$s^\text{spa}_i$ to allow implicit supervision.

\paragraph{Loss Function.}
Given a labeled correct answer $O_j \in \mathbf{O}$, we calculate the cross-entropy loss:
\begin{equation}
\label{eq:lretr}
\small
    \mathcal{L}^\text{RETR} = -\log \frac{\exp(s^\text{spa}_j)}{\sum_{O_i \in \mathbf{O}}\exp(s^\text{spa}_i)} \,.
\end{equation}

Indeed, we employ only QA labels in Eqs.~(\ref{eq:sri})(\ref{eq:lretr}) to implicitly supervise the retriever in Eq.~(\ref{eq:retrieve}) by training the word weighting network in Eq.~(\ref{eq:weighting}).

\subsection{Reader}

Our reader for SQA uses both separate and fused dense representations of paragraphs.

\paragraph{Paragraph Representation (Shared Embeddings).}
Since $\mathbf{P}^\text{top}_i$~contains a small number of $k$~paragraphs, for each paragraph $p_l \in \mathbf{P}^\text{top}_i$ we generate its dense representation capturing its semantic features by reusing the shared PLM in Eq.~(\ref{eq:enc}):
\begin{align}
\label{eq:gil}
\small
\begin{split}
    & [\mathbf{g}_{i,l,1}; \mathbf{g}_{i,l,2}; \ldots] = \\
    & \text{PLM}(\text{[CLS] $p_l$ [SEP] $S$ [SEP] $Q$ [SEP] $O_i$ [SEP]}) \,.
\end{split}
\end{align}
\noindent We take~$\mathbf{g}_{i,l,1}$, the dense representation of the [CLS] token, as an aggregate representation of the entire sequence. We pack together such vectors for all the $k$~paragraphs in~$\mathbf{P}^\text{top}_i$ into a matrix~$\mathbf{G}_i$:
\begin{equation}
\small
    \mathbf{G}_i = [\mathbf{g}_{i,1,1}; \ldots; \mathbf{g}_{i,k,1}] \,.
\end{equation}

\paragraph{Paragraph Re-Scoring.}
For all the $k$~paragraphs in~$\mathbf{P}^\text{top}_i$, we feed their dense representations~$\mathbf{G}_i$ into a dense layer to re-compute their scores~$\mathbf{z}^\text{den}_i$:
\begin{equation}
\label{eq:zid}
\small
    \mathbf{z}^\text{den}_i = [z^\text{den}_{i,1}; \ldots; z^\text{den}_{i,k}]^\intercal \,,~ \text{where }
    z^\text{den}_{i,l} = \text{linear}(\mathbf{g}_{i,l,1}) \,.
\end{equation}
\noindent These scores are expected to be more accurate than those computed by Eq.~(\ref{eq:retrieve}) based on sparse representations. Therefore, let~$\mathbf{P}^\text{fus}_i \subseteq \mathbf{P}^\text{top}_i$ be $k'$~top-ranked paragraphs according to~$\mathbf{z}^\text{den}_i$. Below we will only fuse this subset of paragraphs for both effectiveness and efficiency considerations.

\paragraph{Paragraph Fusion.}
Fusing the knowledge in multiple paragraphs is needed for SQA. We fuse the dense representations of all the $k'$~paragraphs in~$\mathbf{P}^\text{fus}_i$ to enhance their synergy. Specifically, we firstly perform \emph{intra-paragraph fusion}. For each paragraph $p_l \in \mathbf{P}^\text{fus}_i$, we pack together the vectors in Eq.~(\ref{eq:gil}) into four matrices: $\mathbf{G}_{i,l}^\text{P}$ for~$p_l$, $\mathbf{G}_{i,l}^\text{S}$ for~$S$, $\mathbf{G}_{i,l}^\text{Q}$ for~$Q$, and $\mathbf{G}_{i,l}^\text{O}$ for~$O_i$. For each pair of matrices such as~$(\mathbf{G}_{i,l}^\text{P}, \mathbf{G}_{i,l}^\text{S})$, we fuse them into a vector~$\mathbf{f}^\text{PS}_{i,l}$ by dual-attention~\cite{RikiNet}:
\begin{align}
\small
\begin{split}
    (\hat{\mathbf{G}}_{i,l}^\text{P}, \hat{\mathbf{G}}_{i,l}^\text{S}) & = \text{DualAttention}(\mathbf{G}_{i,l}^\text{P}, \mathbf{G}_{i,l}^\text{S}) \,,\\
    \mathbf{f}^\text{P}_{i,l} & = \text{max-pooling}(\hat{\mathbf{G}}_{i,l}^\text{P}) \,,\\
    \mathbf{f}^\text{S}_{i,l} & = \text{max-pooling}(\hat{\mathbf{G}}_{i,l}^\text{S}) \,,\\
    \mathbf{f}^\text{PS}_{i,l} & = \text{relu}(\text{linear}([(\mathbf{f}^\text{P}_{i,l})^\intercal; (\mathbf{f}^\text{S}_{i,l})^\intercal]^\intercal)) \,.
\end{split}
\end{align}
\noindent We concatenate such vectors for all the six pairs of matrices into a vector~$\mathbf{f}_{i,l}$:
\begin{equation}
\small
    \mathbf{f}_{i,l} = [(\mathbf{f}^\text{PS}_{i,l})^\intercal; (\mathbf{f}^\text{PQ}_{i,l})^\intercal; (\mathbf{f}^\text{PO}_{i,l})^\intercal; (\mathbf{f}^\text{SQ}_{i,l})^\intercal; (\mathbf{f}^\text{SO}_{i,l})^\intercal; (\mathbf{f}^\text{QO}_{i,l})^\intercal]^\intercal \,.
\end{equation}
\noindent We pack together such vectors for all the $k'$~paragraphs in~$\mathbf{P}^\text{fus}_i$ into a matrix~$\mathbf{F}_i$:
\begin{equation}
\small
    \mathbf{F}_i = [\mathbf{f}_{i,1}; \ldots; \mathbf{f}_{i,k'}] \,.
\end{equation}
\noindent Then we perform \emph{inter-paragraph fusion} over all the $k'$~paragraphs in~$\mathbf{P}^\text{fus}_i$. We fuse~$\mathbf{F}_i$ into a vector~$\mathbf{f}_i$ by self-attention~\cite{CFC}:
\begin{align}
\small
\begin{split}
    \mathbf{f}_i & = \mathbf{F}_i \times \text{softmax}([a_{i,1}; \ldots; a_{i,k'}]^\intercal) \,,~ \text{where} \\
    a_{i,l} & = \text{tanh}(\text{linear}(\text{tanh}(\text{linear}(\mathbf{f}_{i,l})))) \,.
\end{split}
\end{align}

\paragraph{Option Scoring.}
We leverage both the fused dense representation~$\mathbf{f}_i$ and separate paragraph scores~$\mathbf{z}^\text{den}_i$ to compute two scores for option~$O_i$:
\begin{equation}
\label{eq:sfisdi}
\small
    s^\text{fus}_i = \text{linear}(\mathbf{f}_i) \,,\quad
    s^\text{den}_i = \sum_{p_l \in \mathbf{P}^\text{top}_i}{z^\text{den}_{i,l}} \,.
\end{equation}

\paragraph{Loss Function.}
Given a labeled correct answer $O_j \in \mathbf{O}$, we calculate the cross-entropy loss:
\begin{align}
\label{eq:lread}
\small
\begin{split}
    \mathcal{L}^\text{READ}
    & = -\log \frac{\exp(s^\text{fus}_j)}{\sum_{O_i \in \mathbf{O}}\exp(s^\text{fus}_i)} -\log \frac{\exp(s^\text{den}_j)}{\sum_{O_i \in \mathbf{O}}\exp(s^\text{den}_i)} \\
    & = -\log \frac{\exp(s^\text{fus}_j + s^\text{den}_j)}{\sum_{O_i \in \mathbf{O}}{\sum_{O_{i'} \in \mathbf{O}}{\exp(s^\text{fus}_i + s^\text{den}_{i'})}}} \,.
\end{split}
\end{align}

\subsection{Joint Training and Prediction}

\paragraph{Joint Training.}
We sum the retriever's loss and the reader's loss:
\begin{equation}
\small
    \mathcal{L} = \mathcal{L}^\text{RETR} + \mathcal{L}^\text{READ} \,.
\end{equation}
\noindent We sum such losses over all the training questions.

\paragraph{Prediction.}
For each option $O_i \in \mathbf{O}$, we have computed three scores: $s^\text{spa}_i$, $s^\text{den}_i$, and~$s^\text{fus}_i$. We normalize each score over all the $m$~options by softmax and then calculate their weighted sum as the final score~$s_i$ for~$O_i$:
\begin{align}
\label{eq:si}
\small
\begin{split}
    [\hat{s}^\text{spa}_1; \ldots; \hat{s}^\text{spa}_m]^\intercal & = \text{softmax}([s^\text{spa}_1; \ldots; s^\text{spa}_m]^\intercal) \,,\\
    [\hat{s}^\text{den}_1; \ldots; \hat{s}^\text{den}_m]^\intercal & = \text{softmax}([s^\text{den}_1; \ldots; s^\text{den}_m]^\intercal) \,,\\
    [\hat{s}^\text{fus}_1; \ldots; \hat{s}^\text{fus}_m]^\intercal & = \text{softmax}([s^\text{fus}_1; \ldots; s^\text{fus}_m]^\intercal) \,,\\
    s_i & = \alpha \cdot \hat{s}^\text{spa}_i + \beta \cdot \hat{s}^\text{den}_i + \gamma \cdot \hat{s}^\text{fus}_i \,,
\end{split}
\end{align}
\noindent where $\alpha,\beta,\gamma$ are hyperparameters. We output the option with the highest final score as the answer.
\section{Experiments}
\label{sec:experiments}

All the experiments were performed on RTX 3090.

\subsection{Datasets}

To our knowledge, there were two datasets containing scenario-based multiple-choice ($m=4$) questions, both collected from China's high-school exams. We constructed a new dataset. As shown in Table~\ref{table:datasetstatistics}, scenarios and options are paragraph-long and sentence-long, respectively, thereby challenging existing retrievers as discussed in Section~\ref{sec:introduction}.

\paragraph{GeoSQA}
\cite{GeoSQA} contains geography questions. Each scenario contains a paragraph and a diagram annotated with a description of its main content. We filtered out questions where options are identifiers for objects in diagrams since diagram understanding is outside our research.

\paragraph{GH577}
\cite{gaokao-history} contains history questions. Each scenario contains a paragraph.

\paragraph{GKMC}
standing for \textbf{G}ao\textbf{K}ao-level \textbf{m}ultiple-\textbf{c}hoice questions is a new dataset we constructed containing geography questions. We followed the procedure in \citet{GeoSQA} to crawl and deduplicate questions, but we only kept diagram-free questions to be disjoint from GeoSQA.

\paragraph{Train-Dev-Test Splits.}
We performed five-fold cross-validation with a 3:1:1 split of each dataset into training, development, and test sets.

\subsection{Corpora}
\label{sec:experiments-corpus}

For GeoSQA and GKMC, we combined three geography textbooks and the Chinese Wikipedia into a corpus.
We filtered Wikipedia by only keeping paragraphs in the geography domain identified by a BERT-based classifier fine-tuned with 500~positive and 500~negative paragraphs we manually annotated. The corpus contains 175k~paragraphs.

For GH577 we used all the 5.6m~paragraphs in the Chinese Wikipedia as a corpus.

\begin{table}[t!]
\centering
\small
\resizebox{\columnwidth}{!}{
\begin{tabular}{lrrr}
    \hline
    & GeoSQA & GKMC & GH577 \\
    \hline
    Questions & 3,910 & 1,600 & 577\\
    Chinese characters per scenario & 153 & 82 &  41\\
    Chinese characters per option & 10 & 11 &  15\\
    \hline
\end{tabular}
}
\caption{Dataset statistics.}
\label{table:datasetstatistics}
\end{table}

\subsection{Implementation Details}
\label{sec:experiments-details}

JEEVES builds on a PLM. We implemented two variants of JEEVES based on different PLMs for Chinese: BERT-wwm-ext~\cite{WWM} and ERNIE~\cite{ERNIE}.
We coarse-tuned PLMs on C3 (\url{https://dataset.org/c3/}), a Chinese multiple-choice MRC dataset. We used the BertAdam optimizer. We set $\text{maximum sequence length}=256$, $\text{hidden layer}=12$, $\text{hidden units}=768$, $\text{attention heads}=12$, $\text{dropout rate}=0.1$, $\text{batch size}=16$,
$\text{learning rate}=1e\text{--}5$,
and $\text{warm-up proportion}=0.1$. For GeoSQA and GKMC we set $\text{epochs}=6$ and used two random seeds $\{1,2\}$. For GH577 we set $\text{epochs}=10$ and used four random seeds $\{1,2,3,4\}$.

We used Lucene to implement our retriever.
We equipped Lucene's searcher with a custom scorer using word weights~$\mathbf{w}_i$ to retrieve~$\mathbf{P}^\text{top}_i$.

We tuned six main hyperparameters of JEEVES:
\begin{itemize}
    \item $k \in \{5, 10, 15\}$ below Eq.~(\ref{eq:retrieve}), the number of retrieved paragraphs~$\mathbf{P}^\text{top}_i$;
    \item $k' \in \{2, 3, 4\}$ below Eq.~(\ref{eq:zid}), the number of fused paragraphs~$\mathbf{P}^\text{fus}_i$;
    \item $\tau \in \{100, 200, 300\}$ below Eq.~(\ref{eq:sri}), the number of values in~$\mathbf{z}^\text{spa}_i$ used for computing~$s^\text{spa}_i$;
    \item $\alpha,\beta,\gamma \in \{0, 0.05, \ldots, 1\}$ in Eq.~(\ref{eq:si}), the weights of the three scores for each option.
\end{itemize}

\subsection{Baselines}

We compared with a variety of baseline methods.

We compared with two retriever-reader methods: \textbf{AR}~\cite{AR} and \textbf{DPR}~\cite{DPR}. We adapted the reader in DPR to multiple-choice questions. We coarse-tuned the retrievers in AR (i.e.,~document relevance discriminator) and DPR on DuReader (\url{https://github.com/baidu/DuReader}), a Chinese ODQA dataset containing relevance labels of paragraphs, and we coarse-tuned the readers in AR (i.e.,~answer verifier discriminator) and DPR on C3. We implemented the following variants of DPR using different strategies for distant supervision.
\begin{itemize}
    \item DPR$_\text{exact}$ represents the original implementation in~\citet{DPR}. Paragraphs that contain an exact match of the gold-standard answer were labeled as positives.
    \item DPR$_{\text{appr-}r}$ adopts approximate matching. Paragraphs that contain $\geq$$r\%$~of the non-stopwords in the gold-standard answer were labeled as positives. We set $r \in \{100,50\}$.
    \item DPR$_{\text{silver}}$ adopts an integer linear programming based method for distant supervision~\cite{silver} to label $L$~paragraphs as positives. We set $L=3$ according to the statistics of the development sets.
\end{itemize}
\noindent Table~\ref{table:DS_Datasets} shows the proportions of questions where $\geq$1 positive paragraph was labeled.
Observe that DPR$_\text{exact}$ generated positives for 22.28--40.69\% of the questions in SQA datasets, while it generated positives for 76.68--83.15\% of the questions in ODQA datasets~\cite{DPR}, showing the difficulty of distant supervision for SQA. So we added DPR$_{\text{appr-}r}$ and DPR$_{\text{silver}}$ which generated more positives but might generate false positives.

We compared with a retrieval-based method: \textbf{IR~Solver}~\cite{IR-Solver}. It concatenated the scenario, question, and each option into a query to retrieve the top-ranked paragraph, and used the score of that paragraph as the score of the option.

We compared with five MRC methods: \textbf{SPC}, \textbf{MMM}~\cite{MMM}, \textbf{DCMN+}~\cite{DCMN+}, \textbf{SG-NET}~\cite{SG-Net}, and \textbf{DHC}~\cite{DHC}. We fed them with top-10 paragraphs relevant to each option. We retrieved such paragraphs in a standard way: by concatenating the scenario, question, and option into a query to retrieve using BM25. SPC standing for sentence pair classification employed a sentence pair classifier based on a PLM to rank options by pairing each option with a concatenation of the scenario and the question. For MMM we coarse-tuned its sentence encoder using CMNLI (\url{https://github.com/CLUEbenchmark/CLUE}), a Chinese natural language inference dataset, and we used C3 as its in-domain source dataset for fine-tuning. For SG-NET we used HanLP (\url{https://github.com/hankcs/HanLP}) as its parser.

AR, DPR, SPC, MMM, DCMN+, SG-NET, and DHC build on a PLM. For a fair comparison with JEEVES, we implemented two variants of these methods based on BERT-wwm-ext and ERNIE, configured in the same way as described in Section~\ref{sec:experiments-details}. For readers we tuned maximum sequence length in $\{256,512\}$, epochs in $\{6,10\}$, and learning rate in $\{1e\text{--}5,3e\text{--}5\}$. For the retriever in DPR we tuned maximum sequence length in $\{128,256\}$, set $\text{epochs}=40$, and other hyperparameters according to \citet{DPR}.

More implementation details about the baseline methods are on GitHub.

\begin{table}[t!]
\centering
\small
\begin{tabular}{lrrr}
    \hline
    & GeoSQA & GKMC & GH577 \\
    \hline
    DPR$_\text{exact}$ & 35.68\% & 40.69\% & 22.28\% \\
    DPR$_{\text{appr-}100}$ & 52.43\% & 61.06\% & 33.16\%\\
    DPR$_{\text{appr-}50}$ & 90.18\% & 94.00\% & 79.65\% \\
    DPR$_\text{silver}$ & 100.00\% & 100.00\% & 100.00\% \\
    \hline
\end{tabular}
\caption{Questions with positive paragraphs labeled by different strategies for distant supervision.}
\label{table:DS_Datasets}
\end{table}

\subsection{Evaluation Metrics}

To evaluate SQA, we relied on QA labels to measure the \textbf{accuracy} of each approach: the proportion of correctly answered questions.

To evaluate retrievers for SQA, we sampled 200~questions randomly but evenly from the test set of each fold on GKMC, and for each question we manually annotated relevant paragraphs in the corpus.
We measured the relevance of the paragraphs retrieved by each retriever using
\textbf{MAP} and
\textbf{NDCG} at different positions (@2 and~@10) in the ranked list of retrieved paragraphs. We also measured \textbf{Hit Rate}: the proportion of questions for which $\geq$1 retrieved paragraph is relevant.
\subsection{Results}

We averaged the results over all the random seeds and folds on each dataset.

\begin{table}[t!]
\centering
\small
\resizebox{\columnwidth}{!}{$
\begin{tabular}{lllllll}
\hline
& \multicolumn{2}{c}{GeoSQA} & \multicolumn{2}{c}{GKMC} & \multicolumn{2}{c}{GH577} \\
\cline{2-7}
& dev & test & dev & test & dev & test \\
\hline
IR~Solver & 31.71~$^\bullet$ & 31.71~$^\bullet$ & 45.94~$^\bullet$ & 45.94~$^\bullet$ & 33.33~$^\bullet$ & 33.33~$^\bullet$ \\

\hline
BERT-wwm-ext \\
\quad AR & 34.78~$^\bullet$ & 33.40~$^\bullet$ & 51.06~$^\bullet$ & 50.19~$^\bullet$ & 34.21~$^\bullet$ & 31.75~$^\bullet$ \\
\quad DPR$_\text{exact}$  & 31.20~$^\bullet$ & 29.82~$^\bullet$ & 40.50~$^\bullet$ & 38.86~$^\bullet$ & 34.85~$^\bullet$ & 32.84~$^\bullet$ \\
\quad DPR$_{\text{appr-}100}$  & 32.16~$^\bullet$ & 30.62~$^\bullet$ & 41.08~$^\bullet$ & 39.12~$^\bullet$ & 35.21~$^\bullet$ & 32.84~$^\bullet$ \\
\quad DPR$_{\text{appr-}50}$  & 31.68~$^\bullet$ & 29.87~$^\bullet$ & 41.72~$^\bullet$ & 41.66~$^\bullet$ & 39.41~$^\bullet$ & 32.29~$^\bullet$\\
\quad DPR$_\text{silver}$ &31.46~$^\bullet$ & 31.33~$^\bullet$ & 40.42~$^\bullet$ & 39.06~$^\bullet$& 35.03~$^\bullet$ & 34.85~$^\bullet$\\
\quad SPC  & 36.39~$^\bullet$ & 33.84~$^\bullet$ & 49.81~$^\bullet$ & 48.25~$^\bullet$ & 31.58~$^\bullet$ & 28.07~$^\bullet$ \\
\quad MMM  & 35.96~$^\bullet$ & 32.86~$^\bullet$ & 50.68~$^\bullet$ & 47.69~$^\bullet$ & 40.53~$^\bullet$ & 33.68~$^\bullet$ \\
\quad DCMN+  & 36.19~$^\bullet$ & 34.58~$^\bullet$ & 50.56~$^\bullet$ & 48.38~$^\bullet$ & 37.72~$^\bullet$ & 33.51~$^\bullet$ \\
\quad SG-Net  & 33.76~$^\bullet$ & 33.73~$^\bullet$ & 50.94~$^\bullet$ & 47.69~$^\bullet$ & 40.35~$^\bullet$ & 33.86~$^\bullet$ \\
\quad DHC  & 35.24~$^\bullet$ & 33.84~$^\bullet$ & 48.38~$^\bullet$ & 48.25~$^\bullet$ & 40.35~$^\bullet$ & 32.81~$^\bullet$ \\
\quad JEEVES & \textbf{38.81} & \textbf{36.95} & \textbf{57.34} & \textbf{55.78} & \textbf{46.27} & \textbf{40.35} \\
\hline 
ERNIE \\
\quad AR  & 34.45~$^\bullet$ & 33.30~$^\bullet$ & 49.63~$^\bullet$ & 47.31~$^\bullet$ & 34.39~$^\bullet$ & 32.81~$^\bullet$ \\
\quad DPR$_\text{exact}$  & 32.01~$^\bullet$ & 29.95~$^\bullet$ & 41.70~$^\bullet$ & 39.90~$^\bullet$ & 36.14~$^\bullet$ & 32.46~$^\bullet$\\
\quad DPR$_{\text{appr-}100}$ & 30.75~$^\bullet$ & 30.99~$^\bullet$ & 40.37~$^\bullet$ & 38.06~$^\bullet$ &  36.14~$^\bullet$ & 28.24~$^\bullet$ \\
\quad DPR$_{\text{appr-}50}$  & 31.69~$^\bullet$ & 31.25~$^\bullet$ & 42.50~$^\bullet$ & 40.25~$^\bullet$ & 35.26~$^\bullet$ & 31.23~$^\bullet$\\
\quad DPR$_\text{silver}$ & 32.32~$^\bullet$ & 32.20~$^\bullet$ & 41.56~$^\bullet$ & 40.19~$^\bullet$ & 36.84~$^\bullet$ & 32.51~$^\bullet$\\
\quad SPC  & 33.43~$^\bullet$ & 32.35~$^\bullet$ & 47.63~$^\bullet$ & 46.06~$^\bullet$ & 36.67~$^\bullet$ & 31.76~$^\bullet$ \\
\quad MMM & 34.50~$^\bullet$ & 31.66~$^\bullet$ & 45.88~$^\bullet$ & 40.56~$^\bullet$ & 36.49~$^\bullet$ & 31.23~$^\bullet$ \\
\quad DCMN+  & 34.12~$^\bullet$ & 32.33~$^\bullet$ & 49.06~$^\bullet$ & 47.19~$^\bullet$ & 36.14~$^\bullet$ & 31.58~$^\bullet$ \\
\quad SG-Net  & 32.40~$^\bullet$ & 31.46~$^\bullet$ & 48.57~$^\bullet$ & 46.56~$^\bullet$ & 37.89~$^\bullet$ & 34.21~$^\bullet$ \\
\quad DHC  & 32.84~$^\bullet$ & 33.09~$^\bullet$ & 46.69~$^\bullet$ & 47.31~$^\bullet$ & 36.49~$^\bullet$ & 27.02~$^\bullet$ \\
\quad JEEVES & \textbf{38.76} & \textbf{35.99} & \textbf{57.40} & \textbf{53.78} & \textbf{42.11} & \textbf{36.14} \\
\hline
\end{tabular}
$}
\caption{Comparison with baselines (accuracy of SQA). We mark the results of baselines that are significantly lower than JEEVES under $p<0.01$ ($^\bullet$).}
\label{table:QA_RESULT}
\end{table}


\paragraph{Comparison with Baselines.}
In Table~\ref{table:QA_RESULT}, JEEVES achieved the highest accuracy of SQA on both the validation and test sets of all the three datasets. It significantly outperformed all the baselines under $p<0.01$. The strongest baseline methods were DCMN+ (BERT-wwm-ext) on the test set of GeoSQA, AR (BERT-wwm-ext) on GKMC, and SG-Net (ERNIE) on GH557. JEEVES (BERT-wwm-ext) exceeded them by~2.37 of accuracy on GeoSQA, by~5.59 on GKMC, and by~6.14 on GH577. The results demonstrated the effectiveness of our end-to-end model for SQA.

\begin{table}[t!]
\centering
\resizebox{0.5\columnwidth}{!}{$
\begin{tabular}{lrr}
    \hline
    & BERT-wwm-ext & ERNIE \\
    \hline
    SPC & 102m & 99m \\
    MMM & 109m & 106m \\
    DCMN+ & 109m & 106m \\
    SG-Net & 135m & 133m \\
    DHC & 140m & 137m \\
    \hline
    JEEVES & 130m & 127m \\
    \hline
\end{tabular}
$}
\caption{Comparison with baselines (model size).}
\label{table:parameters_num}
\end{table}

Observe that JEEVES achieved better performance by a model comparable to baseline methods in size. As shown in Table~\ref{table:parameters_num}, their numbers of parameters were of the same order of magnitude.


\paragraph{Comparison between Retrievers.}
In Table~\ref{table:RETRIEVER_RESULT}, JEEVES retrieved the most relevant paragraphs for SQA. It significantly outperformed all the baseline retrievers under $p<0.01$. In particular, JEEVES (BERT-wwm-ext) largely exceeded BM25 by 10.14--11.33 of Hit Rate, 5.36--7.57 of MAP, and 6.88--6.89 of NDCG. Among supervised baseline retrievers, DPR$_{\text{appr-}50}$ was the best-performing variant of DPR but still inferior to AR. Interestingly, none of them were comparable with the unsupervised IR Solver, showing the limitations of existing distant supervision (e.g.,~DPR) and transfer learning (e.g.,~AR) based retrievers for SQA. The results demonstrated the effectiveness of our implicitly supervised retriever for SQA.

\begin{table}[t!]
\centering
\resizebox{\columnwidth}{!}{
\begin{tabular}{lllllll}
\hline
& \multicolumn{2}{c}{Hit Rate} & \multicolumn{2}{c}{MAP}& \multicolumn{2}{c}{NDCG}\\\cline{2-7}
& @2 & @10 & @2 & @10 & @2 & @10\\
\hline
BM25 & 32.76~$^\bullet$ & 56.57~$^\bullet$ & 29.43~$^\bullet$ & 34.11~$^\bullet$ & 24.43~$^\bullet$ & 31.02~$^\bullet$\\
IR~Solver & 37.72~$^\bullet$ & 62.00~$^\bullet$ & 34.00~$^\bullet$ & 38.44~$^\bullet$ & 29.54~$^\bullet$ & 35.39~$^\bullet$\\
\hline
BERT-wwm-ext \\
\quad AR & 37.62~$^\bullet$ & 56.86~$^\bullet$ & 31.95~$^\bullet$ & 35.23~$^\bullet$ & 27.03~$^\bullet$ & 32.13~$^\bullet$\\
\quad DPR$_\text{exact}$ & 25.01~$^\bullet$& 45.57~$^\bullet$& 21.14~$^\bullet$& 23.50~$^\bullet$& 16.15~$^\bullet$& 16.20~$^\bullet$\\
\quad DPR$_{\text{appr-}100}$ & 27.65~$^\bullet$ & 48.38~$^\bullet$ & 23.69~$^\bullet$ & 26.20~$^\bullet$ & 17.92~$^\bullet$ & 17.49~$^\bullet$ \\
\quad DPR$_{\text{appr-}50}$ & 30.80~$^\bullet$& 53.33~$^\bullet$& 26.35~$^\bullet$& 29.04~$^\bullet$& 20.19~$^\bullet$& 20.20~$^\bullet$\\
\quad DPR$_\text{silver}$ & 25.64~$^\bullet$& 44.86~$^\bullet$& 21.43~$^\bullet$& 23.70~$^\bullet$& 16.13~$^\bullet$& 15.67~$^\bullet$\\
\quad JEEVES & \textbf{42.90} & \textbf{67.90} & \textbf{37.00} & \textbf{39.47} & \textbf{31.31} & \textbf{37.91}\\
\hline
ERNIE \\
\quad AR  & 38.85~$^\bullet$ & 60.00~$^\bullet$ & 34.47~$^\bullet$ & 37.39~$^\bullet$ & 29.11~$^\bullet$ & 33.66~$^\bullet$\\
\quad DPR$_\text{exact}$ & 25.46~$^\bullet$& 43.94~$^\bullet$& 21.72~$^\bullet$& 24.06~$^\bullet$& 16.40~$^\bullet$& 15.78~$^\bullet$\\
\quad DPR$_{\text{appr-}100}$ & 25.34~$^\bullet$ & 45.66~$^\bullet$ & 21.91~$^\bullet$ & 24.25~$^\bullet$ & 16.43~$^\bullet$ & 16.25~$^\bullet$\\
\quad DPR$_{\text{appr-}50}$ & 30.27~$^\bullet$& 49.69~$^\bullet$& 26.02~$^\bullet$& 27.95~$^\bullet$& 19.68~$^\bullet$& 18.61~$^\bullet$\\
\quad DPR$_\text{silver}$ & 24.14~$^\bullet$& 44.50~$^\bullet$& 20.85~$^\bullet$& 23.51~$^\bullet$& 15.41~$^\bullet$& 14.90~$^\bullet$\\
\quad JEEVES & \textbf{41.33} & \textbf{65.43} & \textbf{36.83} & \textbf{39.22} & \textbf{31.19} & \textbf{36.94}\\
\hline
\end{tabular}
}
\caption{Comparison between retrievers on the test set of GKMC. We mark the results of baselines that are significantly lower than JEEVES under $p<0.01$ ($^\bullet$).}
\label{table:RETRIEVER_RESULT}
\end{table}

\begin{table}[t!]
\centering
\small
\resizebox{\columnwidth}{!}{
\begin{tabular}{lllllll}
\hline
& \multicolumn{2}{c}{GeoSQA} & \multicolumn{2}{c}{GKMC} & \multicolumn{2}{c}{GH577}\\
\cline{2-7}
& dev & test & dev & test & dev & test\\
\hline
BERT-wwm-ext\\
\quad SPC & 32.30~$^\bullet$ & 31.46~$^\bullet$ & 37.66~$^\bullet$ & 36.83~$^\bullet$ & 37.37~$^\bullet$ & 31.93~$^\bullet$\\
\quad MMM & 35.81~$^\bullet$ & 35.24~$^\circ$ & 52.00~$^\bullet$ & 50.88~$^\bullet$ & 40.70~$^\bullet$ & 35.09~$^\bullet$ \\
\quad DCMN+ & 35.22~$^\bullet$ & 33.17~$^\bullet$ & 50.44~$^\bullet$ & 46.94~$^\bullet$ & 40.35~$^\bullet$ & 37.19~$^\circ$ \\
\quad SG-Net & 34.99~$^\bullet$ & 34.48~$^\bullet$ & 48.25~$^\bullet$ & 47.06~$^\bullet$ & 39.12~$^\bullet$ & 31.05~$^\bullet$ \\
\quad DHC & 34.94~$^\bullet$ & 33.58~$^\bullet$ & 49.13~$^\bullet$ & 48.44~$^\bullet$ & 43.33~$^\circ$ & 38.77 \\
\quad JEEVES & \textbf{38.81} & \textbf{36.95} & \textbf{57.34} &\textbf{55.78} & \textbf{46.27} & \textbf{40.35} \\
\hline
ERNIE\\
\quad SPC & 33.35~$^\bullet$ & 32.22~$^\bullet$ & 39.31~$^\bullet$ & 36.00~$^\bullet$ & 39.30~$^\bullet$ & 31.93~$^\bullet$ \\
\quad MMM &32.51~$^\bullet$ & 32.45~$^\bullet$ & 43.69~$^\bullet$ & 41.31~$^\bullet$ & 35.44~$^\bullet$ & 29.47~$^\bullet$ \\
\quad DCMN+ & 33.63~$^\bullet$ &32.71~$^\bullet$ & 45.19~$^\bullet$ & 43.31~$^\bullet$ & 38.60~$^\bullet$ & 32.28~$^\bullet$ \\
\quad SG-Net &33.07~$^\bullet$ & 32.20~$^\bullet$ & 47.13~$^\bullet$ & 44.50~$^\bullet$ & 38.42~$^\bullet$ & 33.68~$^\bullet$ \\
\quad DHC &33.32~$^\bullet$ & 32.15~$^\bullet$ & 45.19~$^\bullet$ & 44.75~$^\bullet$ & 35.79~$^\bullet$ & 34.56~$^\bullet$ \\
\quad JEEVES & \textbf{38.76} & \textbf{35.99} & \textbf{57.40} & \textbf{53.78} & \textbf{42.11} & \textbf{36.14} \\
\hline
\end{tabular}
}
\caption{Comparison between readers (accuracy of SQA). We mark the results of baselines that are significantly lower than JEEVES under $p<0.01$ ($^\bullet$) or $p<0.05$ ($^\circ$).}
\label{table:READER_RESULT}
\end{table}

\paragraph{Comparison between Readers.}
To compare our reader with baseline readers, we fed the five MRC methods with paragraphs retrieved by our retriever. In Table~\ref{table:READER_RESULT}, JEEVES remained to achieve the highest accuracy of SQA on both the validation and test sets of all the three datasets. It significantly outperformed all the baseline readers under $p<0.01$ or $p<0.05$ except for DHC on the test set of GH577 where the difference was not significant. In particular, JEEVES (BERT-wwm-ext) noticeably exceeded the best-performing baseline reader on the test set of each dataset, i.e.,~MMM (BERT-wwm-ext) on GeoSQA and GKMC, and DHC (BERT-wwm-ext) on GH577, by~1.71, 4.90, and~1.58 of accuracy, respectively. The results demonstrated the effectiveness of our reader for SQA.

\begin{table}[t!]
\centering
\small
\resizebox{\columnwidth}{!}{
\begin{tabular}{lllllll}
\hline
& \multicolumn{2}{c}{GeoSQA} & \multicolumn{2}{c}{GKMC} & \multicolumn{2}{c}{GH577}\\
\cline{2-7}
& dev & test & dev & test & dev & test\\
\hline
BERT-wwm-ext\\
\quad JEEVES & \textbf{38.81} & \textbf{36.95} & \textbf{57.34} &\textbf{55.78} & \textbf{46.27} & \textbf{40.35} \\
\quad w/o $s^\text{den}_i$ & 37.16~$^\bullet$ & 34.44~$^\bullet$ & 54.28~$^\bullet$ & 50.53~$^\bullet$ & 42.72~$^\bullet$ & 36.27~$^\bullet$ \\
\quad w/o $s^\text{fus}_i$ & 37.28~$^\bullet$ & 35.10~$^\bullet$ & 52.75~$^\bullet$ & 49.88~$^\bullet$ & 45.00~$^\circ$ & 38.03 \\
\quad w/o joint training & 37.24~$^\bullet$ & 36.24 & 52.59~$^\bullet$ & 46.96~$^\bullet$ & 45.09~$^\circ$ & 38.60 \\
\hline
ERNIE\\
\quad JEEVES & \textbf{38.76} & \textbf{35.99} & \textbf{57.40} & \textbf{53.78} & \textbf{42.11} & \textbf{36.14} \\
\quad w/o $s^\text{den}_i$ & 37.28~$^\circ$ & 34.30~$^\bullet$ & 53.56~$^\bullet$ & 50.47~$^\bullet$ & 39.39~$^\circ$ & 33.55~$^\circ$ \\
\quad w/o $s^\text{fus}_i$ & 36.89~$^\bullet$ & 35.23 & 53.69~$^\bullet$ & 50.94~$^\bullet$ & 39.52~$^\circ$ & 34.39 \\
\quad w/o joint training & 36.02~$^\bullet$ & 33.68~$^\bullet$ & 51.65~$^\bullet$ & 47.72~$^\bullet$ & 41.92 & 34.21 \\
\hline
\end{tabular}
}
\caption{Ablation study (accuracy of SQA). We mark the results of reduced versions of JEEVES that are significantly lower than the full version of JEEVES under $p<0.01$ ($^\bullet$) or $p<0.05$ ($^\circ$).}
\label{table:ABLATION}
\end{table}

\paragraph{Ablation Study.}
In JEEVES we computed three scores for each option: $s^\text{spa}_i$~in Eq.~(\ref{eq:sri}) based on sparse representations of paragraphs, $s^\text{den}_i$~in Eq.~(\ref{eq:sfisdi}) based on dense representations of separate paragraphs, and $s^\text{fus}_i$~in Eq.~(\ref{eq:sfisdi}) based on fused dense representations of paragraphs. For ablation study we could not remove~$s^\text{spa}_i$ because we relied on it to implicitly supervise our retriever. We removed~$s^\text{den}_i$ and~$s^\text{fus}_i$ to assess their usefulness. In Table~\ref{table:ABLATION}, compared with the full version of JEEVES, the two reduced versions both fell in accuracy on both the validation and test sets of all the three datasets. By removing~$s^\text{den}_i$, the accuracy decreased significantly under $p<0.01$ or $p<0.05$ in all the cases. By removing~$s^\text{fus}_i$, the decreases were also significant in most cases. The results demonstrated the effectiveness of using both separate and fused dense representations of paragraphs for SQA.

In JEEVES we jointly trained our retriever and reader. For ablation study we separately trained them. In Table~\ref{table:ABLATION}, compared with the full version of JEEVES, the reduced version fell in accuracy on both the validation and test sets of all the three datasets. Without joint training, the accuracy decreased significantly under $p<0.01$ or $p<0.05$ in most cases. The results demonstrated the effectiveness of joint training for SQA.

\begin{table}[t!]
\centering
\resizebox{\columnwidth}{!}{
\small
\subfloat[$k$]{
\begin{tabular}{ccc}
\hline
$k$ & dev & test \\
\hline
5 & 55.56 & 52.44\\
10 & \textbf{57.34} & \textbf{55.78} \\
15 & 56.69 & 54.19 \\
\hline
\end{tabular}
\label{table:param_k}
}
\quad
\subfloat[$k'$]{
\begin{tabular}{ccc}
\hline
$k'$ & dev & test \\
\hline
2 & \textbf{57.34} & \textbf{55.78}\\
3 & 57.00 & 54.62 \\
4 & 57.19 & 53.38 \\
\hline
\end{tabular}
\label{table:param_kk}
}
\quad
\subfloat[$\tau$]{
\begin{tabular}{ccc}
\hline
$\tau$ & dev & test \\
\hline
100 & 56.75 & 54.19\\
200 & \textbf{57.34} & \textbf{55.78} \\
300 & 55.37 & 53.75 \\
\hline
\end{tabular}
\label{table:param_tau}
}
}
\caption{Hyperparameter study (accuracy of SQA) of JEEVES (BERT-wwm-ext) on GKMC.}
\label{table:param}
\end{table}

\paragraph{Hyperparameter Study.}
We reported the accuracy of JEEVES (BERT-wwm-ext) on GKMC under different values of its three main hyperparameters: $k \in \{5,10,15\}$, $k' \in \{2,3,4\}$, and $\tau \in \{100,200,300\}$. In Table~\ref{table:param}, the highest accuracy was observed on the validation set under $k=10$, $k'=2$, and $\tau=200$. Therefore, we used these settings throughout all the above experiments.

\begin{table}[t!]
\centering
\small
\begin{tabular}{lrrr}
\hline
Source of Error & GeoSQA & GKMC & GH577 \\
\hline
Corpus & 88\% & 22\% & 14\% \\
Retriever & 8\% & 36\% & 48\% \\
Reader & 14\% & 66\% & 64\% \\
\hline
\end{tabular}
\caption{Error analysis of JEEVES (BERT-wwm-ext).}
\label{table:error}
\end{table}

\paragraph{Error Analysis.}
We randomly sampled 50~questions from the test set of each dataset to which JEEVES (BERT-wwm-ext) outputted an incorrect answer. We analyzed the source of each error. Note that an error could have multiple sources. In Table~\ref{table:error}, corpus was the main source of error on GeoSQA. In particular, commonsense knowledge was needed but often absent in our corpus. Retrieval errors were frequent on GKMC and GH577. Our retriever sometimes assigned large weights to noise words in scenarios and retrieved irrelevant paragraphs. Besides, it could not identify some semantic matches such as \emph{haze} and \emph{air pollution}. Most errors on GKMC and GH577 were related to our reader. Indeed, it could not offer advanced reasoning capabilities such as comparison and negation, which were needed for some questions.

\paragraph{Run Time.}
JEEVES used an average of 0.2~second for answering a question in the test sets.


\section{Conclusion}
\label{sec:conclusion}

Scenario-based multiple-choice questions have posed a great challenge to the retriever-reader framework: keyphrases for retrieving relevant paragraphs are blended with much noise in a long scenario description. In the absence of relevance labels of paragraphs, we devised a novel word weighting mechanism to implicitly train our retriever only using QA labels. It significantly outperformed existing unsupervised, distant supervision based, and transfer learning based retrievers. Based on that, our joint end-to-end model JEEVES exceeded a variety of strong baseline methods on three SQA datasets. While our experiments in the paper are focused on multiple-choice questions in high-school exams, JEEVES has the potential to be adapted to other SQA tasks where scenarios are also very long, such as legal judgment predication. It would be interesting to explore this direction in the future.

Error analysis has revealed some shortcomings of JEEVES, which we will address in future work. Among others, we will incorporate knowledge graphs~\cite{CKGG,DBLP:conf/ccks/ZhangZZHSCLDQ18,DBLP:conf/ijcai/LiH0KG20} and enhance reasoning capabilities~\cite{DBLP:conf/coling/SunCQ18}.

\section*{Acknowledgments}
This work was supported by the National Key R\&D Program of China (2018YFB1005100).

\bibliography{anthology,custom}

\begin{thebibliography}{32}
\expandafter\ifx\csname natexlab\endcsname\relax\def\natexlab#1{#1}\fi

\bibitem[{Chalkidis et~al.(2019)Chalkidis, Androutsopoulos, and
  Aletras}]{DBLP:conf/acl/ChalkidisAA19}
Ilias Chalkidis, Ion Androutsopoulos, and Nikolaos Aletras. 2019.
\newblock \href {https://doi.org/10.18653/v1/p19-1424} {Neural legal judgment
  prediction in {English}}.
\newblock In \emph{{ACL}}, pages 4317--4323.

\bibitem[{Chen et~al.(2017)Chen, Fisch, Weston, and Bordes}]{DRQA}
Danqi Chen, Adam Fisch, Jason Weston, and Antoine Bordes. 2017.
\newblock \href {https://doi.org/10.18653/v1/P17-1171} {Reading {Wikipedia} to
  answer open-domain questions}.
\newblock In \emph{{ACL}}, pages 1870--1879.

\bibitem[{Cheng et~al.(2016)Cheng, Zhu, Wang, Chen, and Qu}]{gaokao-history}
Gong Cheng, Weixi Zhu, Ziwei Wang, Jianghui Chen, and Yuzhong Qu. 2016.
\newblock Taking up the {Gaokao} challenge: An information retrieval approach.
\newblock In \emph{{IJCAI}}, pages 2479--2485.

\bibitem[{Clark et~al.(2016)Clark, Etzioni, Khot, Sabharwal, Tafjord, Turney,
  and Khashabi}]{IR-Solver}
Peter Clark, Oren Etzioni, Tushar Khot, Ashish Sabharwal, Oyvind Tafjord,
  Peter~D. Turney, and Daniel Khashabi. 2016.
\newblock Combining retrieval, statistics, and inference to answer elementary
  science questions.
\newblock In \emph{{AAAI}}, pages 2580--2586.

\bibitem[{Cui et~al.(2019)Cui, Che, Liu, Qin, Yang, Wang, and Hu}]{WWM}
Yiming Cui, Wanxiang Che, Ting Liu, Bing Qin, Ziqing Yang, Shijin Wang, and
  Guoping Hu. 2019.
\newblock Pre-training with whole word masking for {Chinese} {BERT}.
\newblock \emph{CoRR}, abs/1906.08101.

\bibitem[{Huang et~al.(2019)Huang, Shen, Li, Wei, Cheng, Zhou, Dai, and
  Qu}]{GeoSQA}
Zixian Huang, Yulin Shen, Xiao Li, Yuang Wei, Gong Cheng, Lin Zhou, Xinyu Dai,
  and Yuzhong Qu. 2019.
\newblock \href {https://doi.org/10.18653/v1/D19-1597} {{GeoSQA:} {A} benchmark
  for scenario-based question answering in the geography domain at high school
  level}.
\newblock In \emph{{EMNLP-IJCNLP}}, pages 5865--5870.

\bibitem[{Jin et~al.(2020)Jin, Gao, Kao, Chung, and Hakkani{-}T{\"{u}}r}]{MMM}
Di~Jin, Shuyang Gao, Jiun{-}Yu Kao, Tagyoung Chung, and Dilek
  Hakkani{-}T{\"{u}}r. 2020.
\newblock {MMM:} multi-stage multi-task learning for multi-choice reading
  comprehension.
\newblock In \emph{{AAAI-IAAI-EAAI}}, pages 8010--8017.

\bibitem[{Karpukhin et~al.(2020)Karpukhin, Oguz, Min, Lewis, Wu, Edunov, Chen,
  and Yih}]{DPR}
Vladimir Karpukhin, Barlas Oguz, Sewon Min, Patrick S.~H. Lewis, Ledell Wu,
  Sergey Edunov, Danqi Chen, and Wen{-}tau Yih. 2020.
\newblock \href {https://doi.org/10.18653/v1/2020.emnlp-main.550} {Dense
  passage retrieval for open-domain question answering}.
\newblock In \emph{{EMNLP}}, pages 6769--6781.

\bibitem[{Khattab and Zaharia(2020)}]{ColBERT}
Omar Khattab and Matei Zaharia. 2020.
\newblock \href {https://doi.org/10.1145/3397271.3401075} {Colbert: Efficient
  and effective passage search via contextualized late interaction over
  {BERT}}.
\newblock In \emph{{SIGIR}}, pages 39--48.

\bibitem[{Kwiatkowski et~al.(2019)Kwiatkowski, Palomaki, Redfield, Collins,
  Parikh, Alberti, Epstein, Polosukhin, Devlin, Lee, Toutanova, Jones, Kelcey,
  Chang, Dai, Uszkoreit, Le, and Petrov}]{DBLP:journals/tacl/KwiatkowskiPRCP19}
Tom Kwiatkowski, Jennimaria Palomaki, Olivia Redfield, Michael Collins,
  Ankur~P. Parikh, Chris Alberti, Danielle Epstein, Illia Polosukhin, Jacob
  Devlin, Kenton Lee, Kristina Toutanova, Llion Jones, Matthew Kelcey,
  Ming{-}Wei Chang, Andrew~M. Dai, Jakob Uszkoreit, Quoc Le, and Slav Petrov.
  2019.
\newblock {Natural Questions:} a benchmark for question answering research.
\newblock \emph{Trans. Assoc. Comput. Linguistics}, 7:452--466.

\bibitem[{Lai et~al.(2017)Lai, Xie, Liu, Yang, and Hovy}]{RACE}
Guokun Lai, Qizhe Xie, Hanxiao Liu, Yiming Yang, and Eduard~H. Hovy. 2017.
\newblock \href {https://doi.org/10.18653/v1/d17-1082} {{RACE:} large-scale
  reading comprehension dataset from examinations}.
\newblock In \emph{{EMNLP}}, pages 785--794.

\bibitem[{Lally et~al.(2017)Lally, Bagchi, Barborak, Buchanan, Chu{-}Carroll,
  Ferrucci, Glass, Kalyanpur, Mueller, Murdock, Patwardhan, and
  Prager}]{watsonpaths}
Adam Lally, Sugato Bagchi, Michael Barborak, David~W. Buchanan, Jennifer
  Chu{-}Carroll, David~A. Ferrucci, Michael~R. Glass, Aditya Kalyanpur, Erik~T.
  Mueller, J.~William Murdock, Siddharth Patwardhan, and John~M. Prager. 2017.
\newblock {WatsonPaths}: Scenario-based question answering and inference over
  unstructured information.
\newblock \emph{{AI} Magazine}, 38(2):59--76.

\bibitem[{Lee et~al.(2019)Lee, Chang, and Toutanova}]{ORQA}
Kenton Lee, Ming{-}Wei Chang, and Kristina Toutanova. 2019.
\newblock \href {https://doi.org/10.18653/v1/p19-1612} {Latent retrieval for
  weakly supervised open domain question answering}.
\newblock In \emph{{ACL}}, pages 6086--6096.

\bibitem[{Li et~al.(2020)Li, Huang, Cheng, Kharlamov, and
  Gunaratna}]{DBLP:conf/ijcai/LiH0KG20}
Shuxin Li, Zixian Huang, Gong Cheng, Evgeny Kharlamov, and Kalpa Gunaratna.
  2020.
\newblock \href {https://doi.org/10.24963/ijcai.2020/242} {Enriching documents
  with compact, representative, relevant knowledge graphs}.
\newblock In \emph{{IJCAI}}, pages 1748--1754.

\bibitem[{Li et~al.(2021)Li, Sun, and Cheng}]{tsqa}
Xiao Li, Yawei Sun, and Gong Cheng. 2021.
\newblock {TSQA:} tabular scenario based question answering.
\newblock In \emph{{AAAI}}.

\bibitem[{Liu et~al.(2020{\natexlab{a}})Liu, Gong, Fu, Yan, Chen, Jiang, Lv,
  and Duan}]{RikiNet}
Dayiheng Liu, Yeyun Gong, Jie Fu, Yu~Yan, Jiusheng Chen, Daxin Jiang, Jiancheng
  Lv, and Nan Duan. 2020{\natexlab{a}}.
\newblock {RikiNet:} reading {Wikipedia} pages for natural question answering.
\newblock In \emph{{ACL}}, pages 6762--6771.

\bibitem[{Liu et~al.(2020{\natexlab{b}})Liu, Huang, Huang, Liu, and Zhao}]{DHC}
Zhuang Liu, Kaiyu Huang, Degen Huang, Zhuang Liu, and Jun Zhao.
  2020{\natexlab{b}}.
\newblock \href {https://doi.org/10.1145/3340531.3412013} {Dual head-wise
  coattention network for machine comprehension with multiple-choice
  questions}.
\newblock In \emph{{CIKM}}, pages 1015--1024.

\bibitem[{Matsubara et~al.(2020)Matsubara, Vu, and Moschitti}]{rerank}
Yoshitomo Matsubara, Thuy Vu, and Alessandro Moschitti. 2020.
\newblock \href {https://doi.org/10.1145/3397271.3401266} {Reranking for
  efficient {Transformer}-based answer selection}.
\newblock In \emph{{SIGIR}}, pages 1577--1580.

\bibitem[{Pirtoaca et~al.(2019)Pirtoaca, Rebedea, and Ruseti}]{AR}
George{-}Sebastian Pirtoaca, Traian Rebedea, and Stefan Ruseti. 2019.
\newblock \href {https://doi.org/10.18653/v1/D19-1256} {Answering questions by
  learning to rank - learning to rank by answering questions}.
\newblock In \emph{{EMNLP-IJCNLP}}, pages 2531--2540.

\bibitem[{Rajpurkar et~al.(2016)Rajpurkar, Zhang, Lopyrev, and Liang}]{SQuAD}
Pranav Rajpurkar, Jian Zhang, Konstantin Lopyrev, and Percy Liang. 2016.
\newblock \href {https://doi.org/10.18653/v1/d16-1264} {{SQuAD:} 100, 000+
  questions for machine comprehension of text}.
\newblock In \emph{{EMNLP}}, pages 2383--2392.

\bibitem[{Shen et~al.(2021)Shen, Chen, Cheng, and Qu}]{CKGG}
Yulin Shen, Ziheng Chen, Gong Cheng, and Yuzhong Qu. 2021.
\newblock {CKGG:} a {Chinese} knowledge graph for high-school geography
  education and beyond.
\newblock In \emph{{ISWC}}.

\bibitem[{Sun et~al.(2018)Sun, Cheng, and Qu}]{DBLP:conf/coling/SunCQ18}
Yawei Sun, Gong Cheng, and Yuzhong Qu. 2018.
\newblock Reading comprehension with graph-based temporal-casual reasoning.
\newblock In \emph{{COLING}}, pages 806--817.

\bibitem[{Sun et~al.(2019)Sun, Wang, Li, Feng, Chen, Zhang, Tian, Zhu, Tian,
  and Wu}]{ERNIE}
Yu~Sun, Shuohuan Wang, Yu{-}Kun Li, Shikun Feng, Xuyi Chen, Han Zhang, Xin
  Tian, Danxiang Zhu, Hao Tian, and Hua Wu. 2019.
\newblock {ERNIE:} enhanced representation through knowledge integration.
\newblock \emph{CoRR}, abs/1904.09223.

\bibitem[{Wang et~al.(2019)Wang, Yu, Sun, Chen, Yu, McAllester, and
  Roth}]{silver}
Hai Wang, Dian Yu, Kai Sun, Jianshu Chen, Dong Yu, David~A. McAllester, and Dan
  Roth. 2019.
\newblock \href {https://doi.org/10.18653/v1/K19-1065} {Evidence sentence
  extraction for machine reading comprehension}.
\newblock In \emph{{CoNLL}}, pages 696--707.

\bibitem[{Wang et~al.(2018)Wang, Yu, Guo, Wang, Klinger, Zhang, Chang, Tesauro,
  Zhou, and Jiang}]{R3}
Shuohang Wang, Mo~Yu, Xiaoxiao Guo, Zhiguo Wang, Tim Klinger, Wei Zhang, Shiyu
  Chang, Gerry Tesauro, Bowen Zhou, and Jing Jiang. 2018.
\newblock R\({}^{\mbox{3}}\): Reinforced ranker-reader for open-domain question
  answering.
\newblock In \emph{{AAAI-IAAI-EAAI}}, pages 5981--5988.

\bibitem[{Xu et~al.(2020)Xu, Wang, Chen, Pan, Wang, and
  Zhao}]{DBLP:conf/acl/XuWCPWZ20}
Nuo Xu, Pinghui Wang, Long Chen, Li~Pan, Xiaoyan Wang, and Junzhou Zhao. 2020.
\newblock \href {https://doi.org/10.18653/v1/2020.acl-main.280} {Distinguish
  confusing law articles for legal judgment prediction}.
\newblock In \emph{{ACL}}, pages 3086--3095.

\bibitem[{Yang et~al.(2019)Yang, Jia, Zhou, and
  Luo}]{DBLP:conf/ijcai/YangJZL19}
Wenmian Yang, Weijia Jia, Xiaojie Zhou, and Yutao Luo. 2019.
\newblock \href {https://doi.org/10.24963/ijcai.2019/567} {Legal judgment
  prediction via multi-perspective bi-feedback network}.
\newblock In \emph{{IJCAI}}, pages 4085--4091.

\bibitem[{Zhang et~al.(2020{\natexlab{a}})Zhang, Zhao, Wu, Zhang, Zhou, and
  Zhou}]{DCMN+}
Shuailiang Zhang, Hai Zhao, Yuwei Wu, Zhuosheng Zhang, Xi~Zhou, and Xiang Zhou.
  2020{\natexlab{a}}.
\newblock {DCMN+:} dual co-matching network for multi-choice reading
  comprehension.
\newblock In \emph{{AAAI-IAAI-EAAI}}, pages 9563--9570.

\bibitem[{Zhang et~al.(2018)Zhang, Zhang, Zhang, He, Sun, Cheng, Liu, Dai, and
  Qu}]{DBLP:conf/ccks/ZhangZZHSCLDQ18}
Zhiwei Zhang, Lingling Zhang, Hao Zhang, Weizhuo He, Zequn Sun, Gong Cheng,
  Qizhi Liu, Xinyu Dai, and Yuzhong Qu. 2018.
\newblock \href {https://doi.org/10.1007/978-981-13-3146-6\_1} {Towards
  answering geography questions in {Gaokao:} {A} hybrid approach}.
\newblock In \emph{{CCKS}}, pages 1--13.

\bibitem[{Zhang et~al.(2020{\natexlab{b}})Zhang, Wu, Zhou, Duan, Zhao, and
  Wang}]{SG-Net}
Zhuosheng Zhang, Yuwei Wu, Junru Zhou, Sufeng Duan, Hai Zhao, and Rui Wang.
  2020{\natexlab{b}}.
\newblock {SG-Net:} syntax-guided machine reading comprehension.
\newblock In \emph{{AAAI-IAAI-EAAI}}, pages 9636--9643.

\bibitem[{Zhong et~al.(2018)Zhong, Guo, Tu, Xiao, Liu, and Sun}]{legalemnlp18}
Haoxi Zhong, Zhipeng Guo, Cunchao Tu, Chaojun Xiao, Zhiyuan Liu, and Maosong
  Sun. 2018.
\newblock Legal judgment prediction via topological learning.
\newblock In \emph{{EMNLP}}, pages 3540--3549.

\bibitem[{Zhong et~al.(2019)Zhong, Xiong, Keskar, and Socher}]{CFC}
Victor Zhong, Caiming Xiong, Nitish~Shirish Keskar, and Richard Socher. 2019.
\newblock Coarse-grain fine-grain coattention network for multi-evidence
  question answering.
\newblock In \emph{{ICLR}}.

\end{thebibliography}
\bibliographystyle{acl_natbib}

\end{document}